\documentclass[lettersize,journal]{IEEEtran}
\usepackage{amsmath,amsfonts}
\usepackage{algorithmic}
\usepackage{array}
\usepackage[caption=false,font=normalsize,labelfont=sf,textfont=sf]{subfig}
\usepackage{textcomp}
\usepackage{stfloats}
\usepackage{url}
\usepackage{verbatim}
\usepackage{graphicx}
\def\BibTeX{{\rm B\kern-.05em{\sc i\kern-.025em b}\kern-.08em
    T\kern-.1667em\lower.7ex\hbox{E}\kern-.125emX}}
\usepackage{balance}

\usepackage{times}
\usepackage{soul}
\usepackage{url}
\usepackage[hidelinks]{hyperref}
\usepackage[utf8]{inputenc}
\usepackage[small]{caption}
\usepackage{graphicx}
\usepackage{amsmath}
\usepackage{amsthm}
\usepackage{booktabs}
\usepackage{algorithm}
\usepackage{algorithmic}
\usepackage[switch]{lineno}
\usepackage{amsfonts,amssymb}
\usepackage{stfloats}
\usepackage{multirow}
\usepackage{graphicx}
\usepackage{cite}
\usepackage[numbers,sort&compress]{natbib}

\usepackage{colortbl}
\definecolor{light-gray}{gray}{0.82}
\newcolumntype{g}{>{\columncolor{light-gray}}c}
\newcommand*\rot{\rotatebox{75}}

\usepackage{pifont}       
\usepackage{bbding}       
\usepackage{fontawesome}  

\usepackage{makecell}
 
\newcommand{\xmark}{\ding{55}}

\begin{document}
\title{How Noise Benefits AI-generated Image Detection}
\author{ Ziqiang Li, Jiazhen Yan, Fan Wang, Kai Zeng, Ziwen He, Zhangjie Fu,~\IEEEmembership{Member,~IEEE,}
\thanks{This work was supported in part by the National Natural Science Foundation of China under grant U22B2062, 62172232, and Jiangsu Provincial Science and Technology Major Project (No. BG2024042). (Corresponding author: Zhangjie Fu).}
\thanks{Jiazhen Yan, Ziqiang Li, Ziwen He and Zhangjie Fu are with the Engineering Research Center of Digital Forensics, Ministry of Education, Nanjing University of Information Science and Technology, Nanjing, 210044, China. (e-mail: 247918horizon@gmail.com, iceli@mail.ustc.edu.cn, \{ziwen.he, fzj\}@nuist.edu.cn).}
\thanks{Fan Wang is with the University of Macau, Macau, 999078, China. (e-mail: wf71103@126.com.}
\thanks{Kai Zeng is with the University of Siena, Siena, Italy. (e-mail: kai.zeng@unisi.it.}
}

\markboth{Journal of \LaTeX\ Class Files,~Vol.~18, No.~9, September~2020}%
{How to Use the IEEEtran \LaTeX \ Templates}

\maketitle

\begin{abstract}


Generalization to unseen generative models remains a fundamental challenge in AI-generated image (AIGI) detection. Despite recent advances, existing detectors often overfit to spurious shortcuts—such as image-type biases, semantic correlations, and frequency-domain discrepancies—that emerge early during optimization and fail to capture causal forensic cues. This shortcut reliance leads to rapid training loss convergence but poor cross-domain robustness under open-world conditions.
In this paper, we present a noise-driven learning framework for AIGI detection. We first reveal that injecting subtle feature-space perturbations into pretrained encoders can effectively mitigate early-stage shortcut domination and stabilize decision margins. Building on this insight, we propose PiND (Positive-incentive Noise for AI-generated image Detection), a unified variational framework that jointly learns a task-adaptive, forgery-conditional noise generator and a detection network. Unlike random noise, PiND optimizes structured Gaussian perturbations to maximize mutual information with the detection task, thereby reducing conditional uncertainty and suppressing shortcut-sensitive directions while preserving stable artifact-related cues.
Extensive experiments on two standard benchmarks and three real-world degradation benchmarks demonstrate that PiND consistently improves robustness and generalization. Our method achieves new state-of-the-art performance, delivering up to 8.0\% absolute gains in average accuracy over existing approaches.

\end{abstract}

\begin{IEEEkeywords}
AI-generated image detection, AI security,  Positive-incentive Noise.
\end{IEEEkeywords}

\section{Introduction}
\label{sec:intro}

\IEEEPARstart{W}{ith} the rapid advancement of generative models, including Generative Adversarial Networks (GANs)~\cite{li2023systematic,li2022fakeclr,karras2019style} and diffusion-based architectures~\cite{wu2024infinite,rombach2022high,croitoru2023diffusion}, the boundary between synthetic and real visual content has become increasingly blurred. While these models enable compelling applications, they also amplify risks such as misinformation, privacy violations, and challenges to digital forensics. Consequently, AI-generated image (AIGI) detection has become a critical task. Despite notable progress, current detectors often suffer substantial performance degradation on out-of-distribution (OOD) samples, \textit{e.g.}, images generated by unseen model architectures, novel diffusion pipelines, or diverse post-processing operations, thereby limiting their reliability in real-world forensic settings.

\begin{figure}
    \centering
    \includegraphics[width=1\linewidth]{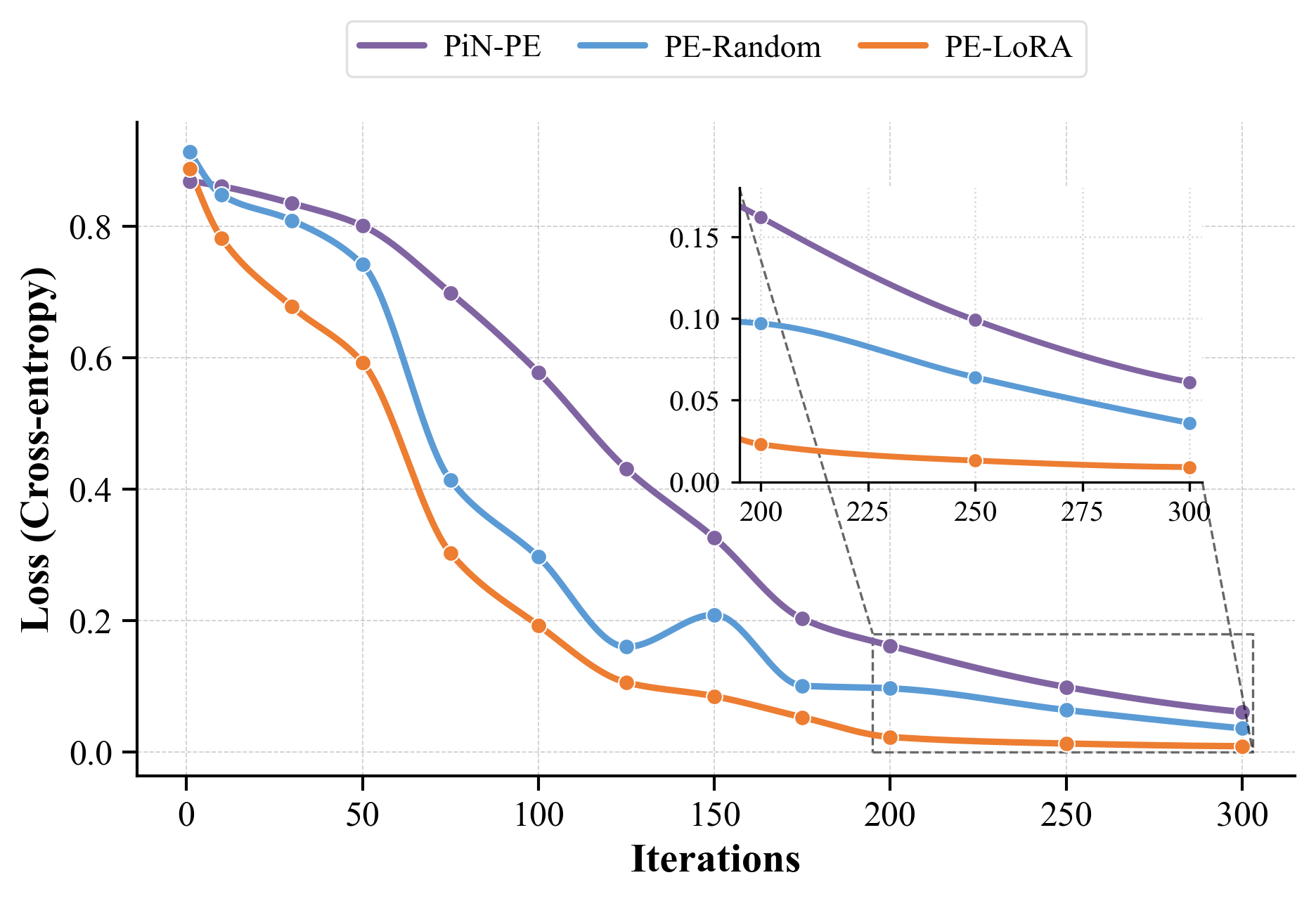}
    \caption{\textbf{The Cross-entropy Loss during Training.} When training the PE-LoRA network, the cross-entropy loss drops sharply to approximately 0.2 within about 100 iterations, indicating early overfitting. Introducing tiny feature-space perturbations (PE-Random) mitigates this effect to some extent, maintaining a higher loss of around 0.3 over the same iterations. However, he benefit of random perturbations is unstable and may disrupt optimization, as evidenced by a sudden loss increase after roughly 150 iterations. In contrast, our proposed PiND both slows the initial loss decrease and maintains more stable margins throughout training, consistent with suppressing shortcut-sensitive directions and promoting more robust convergence.}
    \label{fig:introduction}
\end{figure}

To address these challenges, existing methods can be broadly categorized into two paradigms: \textbf{i) Invariance Learning}, which aims to extract task-relevant cues through contrastive objectives~\cite{koutlis2024leveraging,tan2025c2p,baraldi2025contrasting,liang2025transfer} or handcrafted priors, such as local pixel inconsistencies~\cite{chen2020manipulated,tan2024rethinking,zheng2024breaking}, frequency-domain artifacts~\cite{qian2020thinking,luo2021generalizing,tan2024frequency,yan2026dual}, and category-interference patterns~\cite{yan2025ns,zhang2025towards,tao2025sagnet}; and \textbf{ii) Data Diversification}, which mitigates overfitting by expanding training coverage via cross-generator datasets~\cite{guillaro2025bias,yang2025d,zhou2025breaking} or data augmentation strategies~\cite{wang2020cnn,chen2024drct,li2025improving}. However, under open-world settings, the training distribution can never exhaustively encompass all possible generative models and post-hoc manipulations. Recent studies further indicate that standard training pipelines tend to induce \emph{spurious shortcuts}, such as image-type biases~\cite{grommelt2024fake,guillaro2025bias}, semantic correlations~\cite{guillaro2025bias,zhang2025towards,yan2025ns}, and frequency-domain discrepancies~\cite{chen2025dual}, and have proposed targeted strategies to suppress their influence. Despite these efforts, we observe that detectors still tend to overfit to previously unseen \textit{spurious shortcuts} present in the training data rather than genuine artifact-related cues, resulting in early overfitting and poor generalization. This behavior arises because shortcut features often emerge rapidly during optimization and are difficult to eliminate completely. As illustrated in Fig.~\ref{fig:introduction}, the training loss drops below 0.1 within only a few iterations, suggesting that shortcut learning occurs at an early stage and constitutes a key factor underlying the observed generalization gap.

Motivated by these observations, we investigate whether \emph{tiny perturbations in feature space} can steer learning away from shortcut reliance and toward causal forensic cues. Specifically, we inject small zero-mean Gaussian noise into the feature embeddings of the image encoder. This simple intervention yields a pronounced effect: early-stage shortcut domination is alleviated, and decision margins stabilize for a subset of samples (Fig.~\ref{fig:introduction}). Remarkably, despite its simplicity, this strategy already delivers substantial performance gains on unseen generators (see Sec.~\ref{sec:ablation}). These findings point to a key insight: \textbf{tiny feature-space perturbations can disproportionately disrupt brittle shortcut evidence, slowing its premature dominance while preserving margins supported by stable, artifact-related cues.}

Nevertheless, random noise is not task-aligned for AIGI detection. This motivates a key question: \textbf{Can we design customized noise to fine-tune a pretrained model more effectively?} Inspired by the variational \emph{Positive-Incentive Noise} principle~\cite{zhang2025variational,huang2025enhance,li2022positive}, which demonstrates that appropriately structured noise can reduce conditional task uncertainty and enhance generalizability, we propose \textbf{PiND}—\underline{P}ositive-\underline{i}ncentive \underline{N}oise for AI-generated image \underline{D}etection. PiND elevates random jitter into a controllable and optimizable training signal by jointly learning task-anchored, artifact-conditional Gaussian perturbations alongside the detector. Specifically, we construct a task-relevant perturbation $\mathcal{E}$ that can decrease the uncertainty of predictions and simplify the task, which is expressed as:
\begin{equation}
    I(\mathcal{T}; \mathcal{E}) > 0 \; \Leftrightarrow \; H(\mathcal{T}) > H(\mathcal{T} | \mathcal{E}),
\end{equation}
where $\mathcal{T}$ is the definition of the AIGI detection task from a probabilistic perspective, $I(\cdot; \cdot)$ denotes mutual information, and H represents entropy. We design a lightweight cross-modal attention module that integrates visual representations with label embeddings to produce a \emph{forgery-aware, curvature-regularized} perturbation, and optimize it to maximize the mutual information $I(\mathcal{T};\mathcal{E})$ between the injected noise $\mathcal{E}$ and the detection task $\mathcal{T}$. Intuitively, we hypothesize that images generated by each generative model follow a distinct distribution, which can be counteracted by introducing task-specific perturbations from a tailored distribution. This effect is further amplified by optimizing the noise itself. In contrast to random noise, which can negatively impact performance by artificially increasing input complexity, our task-related perturbation simplifies the task and suppresses spurious shortcuts, ultimately enhancing the model's generalization ability.

Under these structured yet stochastic transformations, the shared encoder is encouraged to suppress shortcut-sensitive components while reinforcing stable forensic cues, leading to more consistent and trustworthy predictions. What's more, our method is not tied to any particular pretrained model and can be applied broadly to ViT-based pretrained architectures.

Our main contributions can be summarized as follows:
\begin{itemize}

\item We reveal that introducing subtle, feature-space perturbations can effectively steer model learning away from spurious shortcuts and toward causal, generalizable cues, providing a principled perspective on noise-driven regularization in AIGI detection.

\item  We propose a unified variational training framework that jointly learns a $\pi$-noise generator and a detection network, enabling task-adaptive, forgery-conditional noise injection guided by the optimization objective. This mechanism suppresses shortcut-sensitive directions while amplifying stable, task-relevant forensic evidence under beneficial stochastic transformations. 
    
\item Extensive experiments across multiple generative domains and unseen architectures show that our method consistently enhances cross-domain robustness and generalization, establishing a new paradigm for noise-driven learning in AIGI detection.

\end{itemize}

\section{Related Work}

\subsection{Towards Generalizable AI-generated Image Detection}
As generative model technology continues to advance rapidly, ensuring the accurate identification of synthetic images has become increasingly critical. Much research has been devoted to generalizable AI-generated image detection, which can be broadly categorized into two categories: invariance learning and data diversification. 

\noindent\textbf{Invariance Learning.} Many methods are dedicated to extracting task-invariant representations to improve the generalization ability of the model, including frequency-domain analysis \cite{qian2020thinking,luo2021generalizing,tan2024frequency,yan2026dual}, local pixel differences \cite{chen2020manipulated,cavia2024real,tan2024rethinking,zheng2024breaking}, gradient patterns \cite{tan2023learning}, reconstruction errors \cite{wang2023dire,cazenavette2024fakeinversion,chu2025fire}, and contrastive learning \cite{liang2025transfer}. For instance, FreqNet \cite{tan2024frequency} utilizes the Fast Fourier Transform (FFT) to extract global high-frequency information; LGrad \cite{tan2023learning} leverages gradients from pretrained CNNs to reveal universal forgery traces; DIRE \cite{wang2023dire} extracts the reconstruction error of the image over a pre-trained diffusion model as the artifact of diffusion-based images; NTF \cite{liang2025transfer} significantly improves the generalization ability of forged image detection by utilizing natural trace representation learning and soft contrastive learning for pre-training. In addition, benefiting from large-scale pre-training, many recent methods have begun to leverage vision–language models (VLMs) (such as CLIP \cite{radford2021learning}) for AI-generated image detection, including feature-based \cite{ojha2023towards,koutlis2024leveraging,zhang2025towards} and fine-tuning-based  \cite{liu2024forgery,yan2024orthogonal,tan2025c2p,yan2025ns} methods. To illustrate, UnivFD \cite{ojha2023towards} directly freezes CLIP's visual encoder and tunes only a linear layer for classification; Effort \cite{yan2024orthogonal} employs Singular Value Decomposition to construct two orthogonal subspaces to preserve the pre-trained knowledge while learning forgery-related patterns; C2P-CLIP \cite{tan2025c2p} finetunes the image encoder by constructing category concepts combined with contrastive learning; NS-Net \cite{yan2025ns} constructs a NULL-Space of semantic features to remove semantic interference embedded in visual features. Despite notable progress, existing detectors still tend to exploit spurious shortcuts present in the training data rather than learning the causal mechanisms underlying artifact formation, which remains the core issue limiting their generalization to unseen generative models.

\noindent\textbf{Data Diversification.} Some studies adopt a data-driven perspective, aiming to enhance artifact visibility through data diversification, thereby improving the model’s generalization ability, which can be broadly categorized into data augmentation \cite{wang2020cnn,chen2024drct,li2025improving} and dataset construction \cite{yang2025d,zhou2025breaking}. Specifically, CNN-Spot \cite{wang2020cnn} employs diverse data augmentation techniques to enhance generalization to unseen testing data; SAFE \cite{li2025improving} integrates cropping and augmentations such as ColorJitter, RandomRotation to improve generalization. What's more, DRCT \cite{chen2024drct} utilizes reconstructed images as informative yet challenging samples, enabling the detector to learn subtle distinctions between real and generated images; OMAT \cite{zhou2025breaking} significantly enhances the model’s generalization and adversarial robustness by incorporating optimized adversarial samples into the training set. However, in open-world environments, the training distribution can never fully encompass all possible generators. Therefore, data diversification alone cannot fundamentally resolve the problem of spurious shortcuts, which ultimately limits the generalization capability of the detector.

\subsection{Shortcuts Existing in AI-generated Image Detection}
Recent studies on AI-generated image detection suggest that common training pipelines may induce spurious shortcuts, such as image-type cues~\cite{grommelt2024fake,guillaro2025bias}, semantic content biases~\cite{guillaro2025bias,zhang2025towards,yan2025ns}, and frequency-domain discrepancies~\cite{chen2025dual}. To illustrate, Grommelt \textit{et al.} identify the dataset-level biases related to JPEG compression and image resolution, and emphasize that detectors should not inadvertently rely on such undesirable variables. B-Free \cite{guillaro2025bias} further analyzes semantic-content bias and constructs debiased training samples via reconstruction, encouraging detectors to focus on generation-related artifacts rather than content. In addition, DDA \cite{chen2025dual} shows that pixel-level alignment alone cannot fully bridge the gap between real and synthetic images, and proposes a dataset construction strategy that aligns pairs in both pixel and frequency domains to mitigate frequency-related bias. Despite these effective suppressions of known biases, detectors can still overfit to unknown spurious shortcuts in the training data instead of genuine artifact-related cues, resulting in early overfitting and poor generalization (as shown in Figure~\ref{fig:introduction}).

\subsection{Positive Noise}
In fact, within the signal-processing society, it has been demonstrated that random noise helps stochastic resonance improve the detection of weak signals \cite{benzi1981mechanism}. Noise can have a positive effect when the mixing probability distribution lies outside the extreme region \cite{sethna2001crackling}. It has also been reported that noise can enhance model generalization in natural language processing (NLP) tasks \cite{xie2017data,pereira2021multi,aghajanyan2020better,jiang2020smart}. Recently, the Positive-incentive Noise \cite{li2022positive} has been proposed to scientifically investigate the positive or negative impact brought by noise. Similarly, other methods \cite{zhang2024data,huang2025enhance,zhang2025variational} extended this insight to contrastive learning and vision-language alignment.

To the best of our knowledge, we are among the first to systematically study how feature-space perturbations affect the generalization of AI-generated image detectors. As discussed in Sec. \ref{sec:intro}, small perturbations in the feature space can mitigate the effects of spurious shortcuts and improve generalization. To make such perturbations inherently task-oriented, we construct task-relevant noise with artificial-level guidance from text embeddings. This transforms the temporary random perturbations into predictable, artifact-conditional signals, suppressing directions sensitive to shortcuts while amplifying stable forensic clues. Fundamentally different from existing methods, which are based on generalization artifact representation and data diversification, our approach explores the potential of noise in generated image detection from a novel perspective and achieves superior performance.

\begin{figure*}[!t]
    \centering
    \includegraphics[width=1\linewidth]{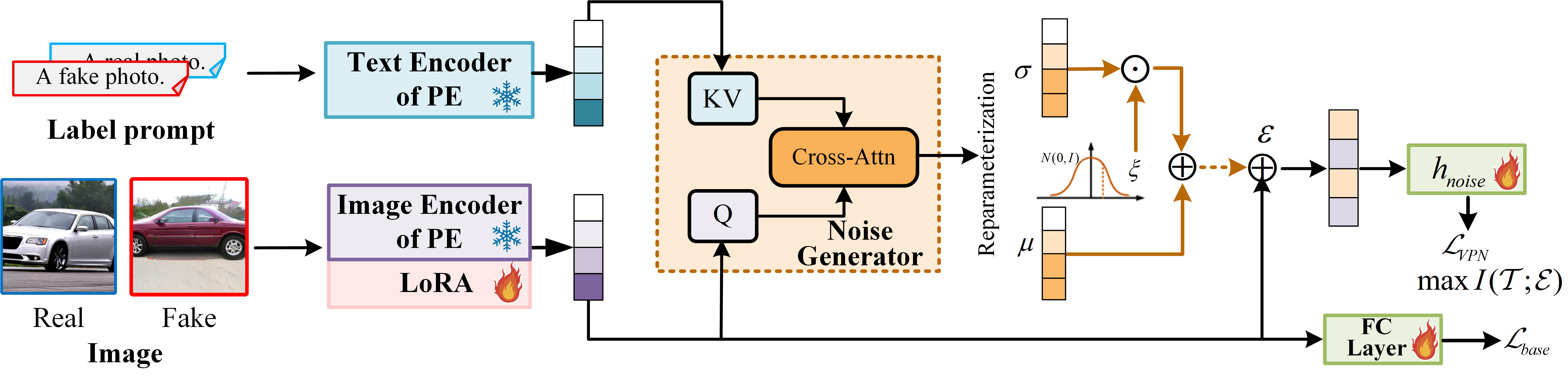}
    \caption{\textbf{Architecture of PiND for Generalizable AI-Generated Image Detection}. It includes a conditional noise generator and a detection network, enabling task-adaptive, forgery-conditional noise injection guided by the optimization objective. This mechanism suppresses shortcut-sensitive directions while amplifying stable, task-relevant forensic evidence under beneficial stochastic transformations. }
    \label{fig:backbone}
\end{figure*}

\section{Methodology}

Generalization to out-of-distribution samples remains a fundamental challenge in AI-generated image detection. We identify spurious shortcuts as the primary cause of overfitting in existing detectors. To address this issue, we introduce Positive-incentive Noise to explore the constructive role of noise in AI-generated image detection from a novel perspective, termed Positive-incentive Noise for AI-generated image detection (PiND).

In this section, we first provide the problem formulation in Section \ref{sec:3.1}, and then introduce how to define the task entropy on the specific dataset $\mathcal{X}$, which is vital for calculating the optimization objective $I(\mathcal{T}; \mathcal{E})$ when generating $\pi$-noise. To optimize this objective, the variational approximation is applied to obtain its upper bound in Section \ref{sec:3.2}. In the final approximate loss, we define the joint training process with $\pi$-noise, which includes both the architecture of the noise generator and the parameter updates during training, as discussed in Section \ref{sec:3.3}.


\subsection{Problem Formulation}
\label{sec:3.1}

Given an image $x \in \mathcal{X}$ and its binary label $y \in \mathcal{Y} = \{0, 1\}$ indicating whether it is \emph{real} or \emph{fake}, 
the goal of AIGI detection is to learn a detector with classifier $h_\phi : \mathcal{X} \rightarrow \mathcal{Y}$ that estimates the posterior probability, which can be expressed as:
\begin{equation}
q_\phi(y | x) = \operatorname{softmax}(\hat{h}_\phi(x)).
\label{eq:task}
\end{equation}

To quantify the intrinsic difficulty of the detection task on an arbitrary dataset, we first measure its \emph{task entropy}. Following prior work~\cite{zhang2025variational,huang2025enhance,li2022positive}, task entropy can be used to formulate the complexity as
\begin{equation}
\begin{aligned}
H(\mathcal{T}) = H(y | x)  &=\mathbb{E}_{x \sim {\mathcal{X}}} \mathbb{E}_{y \sim p(y | x)}[-\log p(y | x)] \\
& =\mathbb{E}_{p(x, y)}[-\log p(y | x)],
\end{aligned}
\end{equation}
which characterizes the uncertainty of predicting the authenticity label. However, in real-world environments, the distributions of $\mathcal{X}$ generated by different generation models vary, which increases the uncertainty of predicting the label $y$.

As discussed in Sec \ref{sec:intro}, we consider introducing the task-related noise $\mathcal{E}$ into the detection task. Similar to $H(\mathcal{T})$, the conditional task entropy under the injected noise $\mathcal{E}$ can be defined as:
\begin{equation}
H(\mathcal T|\mathcal E)=\mathbb{E}_{p(x, y,\varepsilon)}[-\log p(y | x,\varepsilon)].
\end{equation}
Then, the mutual information between the task and the injected noise $I(\mathcal{T}; \mathcal{E})$ is defined as
\begin{equation}
I(\mathcal{T}; \mathcal{E}) = H(\mathcal{T}) - H(\mathcal{T} | \mathcal{E}).
\label{eq:multual}
\end{equation}
Thus, the noise $\mathcal{E}$ can decrease the uncertainty of predictions and simplify the task if it satisfies the following conditions:
\begin{equation}
    I(\mathcal{T}; \mathcal{E}) > 0 \; \Leftrightarrow \; H(\mathcal{T}) > H(\mathcal{T} | \mathcal{E}).
\end{equation}
Therefore, our core objective is to maximize the mutual information $I(\mathcal{T}; \mathcal{E})$ to improve the model's ability to distinguish between real and fake images. Since $H(\mathcal{T})$ is constant during optimization, maximizing $I(\mathcal{T}; \mathcal{E})$ is equivalent to minimizing the conditional task entropy $H(\mathcal{T} | \mathcal{E})$.
This encourages the detector to learn noise-invariant yet task-discriminative representations, thereby improving its robustness and generalization to unseen generative models and domains.

\subsection{Variational Objective}
\label{sec:3.2}

The variational objective of our framework is derived from the non-negativity property of the Kullback–Leibler (KL) divergence, expressed as:

\begin{equation}
\mathrm{KL}(p \| q) \geq 0 
\;\Leftrightarrow\;
\mathbb{E}_{p(x)}[\log p(x)] 
\geq 
\mathbb{E}_{p(x)}[\log q(x)].
\end{equation}
Based on this property, we apply the principle of variational inference to obtain a tractable upper bound of the conditional task entropy $H(\mathcal{T} | \mathcal{E})$:
\begin{equation}
\mathcal{L}_\text{VPN} 
= \mathbb{E}_{p(x, y, \varepsilon)}[-\log q(y | x, \varepsilon)] 
\geq 
H(\mathcal{T} | \mathcal{E}),
\end{equation}
where $q(y | x, \varepsilon)$ is a variational approximation of the intractable posterior $p(y | x, \varepsilon)$.  
Intuitively, minimizing $\mathcal{L}_\text{VPN}$ reduces the conditional uncertainty of task predictions, thereby guiding the detector toward more confident and reliable decisions.
To avoid direct integration over the continuous data distribution, we approximate the expectation in $\mathcal{L}_\text{VPN}$ using Monte Carlo sampling:
\begin{equation}
\mathcal{L}_\text{VPN}
\approx 
\frac{1}{n} 
\sum_{i=1}^n 
\mathbb{E}_{p(\varepsilon | x_i, y_i)}
\left[-\log q(y_i | x_i, \varepsilon)\right].
\end{equation}
We assume the conditional noise distribution follows a Gaussian form,
\begin{equation}
p(\varepsilon | x_i, y_i) 
= \mathcal{N}(\mu_\theta(x_i, y_i), \Sigma_\theta(x_i, y_i)),
\end{equation}
where the mean $\mu_\theta$ and covariance $\Sigma_\theta$ are outputs of a learnable function $f_\theta(x_i, y_i)$ parameterized by $\theta$.  
Since the integral over the continuous variable $\varepsilon$ remains intractable, we further employ the \emph{reparameterization trick}~\cite{kingma2013auto} to enable gradient backpropagation through stochastic nodes.  
Specifically, we express the sampling process as:
\begin{equation}
\varepsilon 
= G_\theta(x_i, y_i, \epsilon) 
= \Sigma_\theta(x_i, y_i) \cdot \epsilon 
+ \mu_\theta(x_i, y_i),
\label{eq:noise}
\end{equation}
where $\epsilon \sim \mathcal{N}(0, I)$ is a standard Gaussian random variable, and $G_\theta$ denotes a differentiable function that generates task-oriented noise.

Under the Monte Carlo approximation, the final training objective can be written as:
\begin{equation}
\begin{aligned}
\mathcal{L}_\text{VPN} 
&\approx 
\frac{1}{n} 
\sum_{i=1}^n 
\mathbb{E}_{p(\epsilon)}
\left[
-\log q\big(y_i | x_i, G_\theta(\epsilon, x_i, y_i)\big)
\right] \\
&\approx 
\frac{1}{n \cdot m} 
\sum_{i=1}^n 
\sum_{j=1}^m 
\Big[
-\log q\big(y_i | x_i, 
G_\theta(\epsilon_{ij}, x_i, y_i)\big)
\Big],
\end{aligned}
\label{eq:vpn}
\end{equation}
where $m$ denotes the number of noise samples per input.  
Accordingly, the final training object contains two components:
\begin{equation}
\mathcal{L}=\mathcal{L}_\text{base}+\gamma\mathcal{L}_\text{VPN},
\label{eq:loss}
\end{equation}
where $\gamma$ is a trade-off coefficient and $\mathcal{L}_\text{base}$ represents the original cross entropy loss of the base model $h_\phi$.


\subsection{Joint Training with $\pi$-noise}
\label{sec:3.3}
The formula Eq. (\ref{eq:vpn}) derived above is mainly divided into two parts: \textbf{I)} $q\big(y | x, \mathcal{E}\big)$, which models label prediction under the injected noise. We will discuss it below in detail. \textbf{II)} $G_\theta(\epsilon, x, y)$, which generates task-related noise. Different from the common practice of injecting noise at the image level, we apply perturbations directly in the feature space. Concretely, we replace the raw image $x$ with the visual feature $f$ extracted by the visual encoder, and replace the label $y$ with the text feature $t$ extracted by the text encoder, and generate task-relevant perturbations $\mathcal{E}$ conditioned on these representations. By optimizing the generator $G_\theta(\cdot)$ via Eq.~(\ref{eq:loss}) in a joint training framework, the visual encoder is encouraged to learn more stable and generalizable artifact representations.

Leveraging the powerful generalization capability of pretrained models in AI-generated image detection tasks, we select the pre-trained PE (Perception Encoder) model as the base model. The framework has been illustrated in Figure \ref{fig:backbone}. It is worth mentioning that our method is not limited to a single pre-trained model; it is applicable to multiple ViT-based pre-trained models simultaneously, as shown in Table \ref{tab:backbone}.

\noindent\textbf{Input and Feature Extraction.}
For each image $x$, we attach a minimal text prompt $z$ that is aligned with its label, such as "A real photo." for real images, and "A fake photo." for fake images. The PE model provides an image encoder $E_{\text{img}}$ and a text encoder $E_{\text{text}}$, the feature extraction processes can be defined as:
\begin{equation}
    f \;=\; E_{\text{img}}(x;\varphi )\in\mathbb{R}^d,\;\;\;\; t \;=\; E_{\text{text}}(z;\phi)\in\mathbb{R}^d,
\end{equation}
where $\varphi$ denotes the trainable parameters of the image encoder, and $\phi$ is frozen.

\noindent\textbf{Architecture of Noise Generator.}
As illustrated in Eq. (\ref{eq:noise}), generating conditional noise typically relies on explicit label information. Instead of using conventional one-hot labels, we condition the perturbation on label-aligned textual embeddings $t$ extracted from PE’s text encoder, enabling the noise generator to adapt its perturbations more precisely to the underlying forensic categories. Given the extracted image feature $f \in \mathbb{R}^d$ and the corresponding text feature $t \in \mathbb{R}^d$, we instantiate a lightweight cross-attention module $G_\theta(f, t)$ to generate per-dimension Gaussian parameters $(\mu, \text{var})$ in the feature space instead of the image space. Specifically, we define the weight matrices $W_q,W_k,W_v,W_\mu,W_{\text{var}}\in\mathbb{R}^{d\times H}$, where $H=d/r$ is a bottleneck hidden size ($r{\ge}1$). The generation process can be formulated as:
\begin{gather}
q=W_q f,\quad k=W_k t,\quad v=W_v t,\nonumber \\ 
a=\text{SoftMax}(q k^\top)v\in\mathbb{R}^{H},\\
\mu=W_\mu a,\qquad \text{var}=\psi\!\left(W_{\text{var}} a\right), \nonumber
\end{gather}
where $\psi(\cdot)$ enforces element-wise nonnegativity. Reparameterized sampling and feature perturbation are as follows:
\begin{equation}
\varepsilon=\mu+\text{var}\odot \xi,\quad \xi\sim\mathcal{N}(0,I_d),\qquad 
\tilde f=f+\varepsilon.
\end{equation}

In this way, we combine the artifact-conditional Gaussian perturbations $\varepsilon$ with artifact representation $f$ to maximize $I(\mathcal{T}; \mathcal{E})$ as illustrated in (\ref{eq:multual}).

\begin{algorithm}[t]
\caption{\textbf{Training Procedure of PiND}}
\label{alg:pin-clip-train}
\small
\noindent\textbf{Input:}
Traing set  $\mathcal{D} = \{x_j, y_j\}_{j=1}^N,$ with text prompts $z_i$, learning rate $\mu$, hyperparameter $\lambda$.\\
\textbf{Output:}
Trained model parameters $\varphi$, $\theta$, $h$, and $h_{\text{noise}}$.

\noindent\textbf{Step 1: Feature Extraction}\\
Extract features $ f_j = E_{\text{img}}(x_j;\varphi), t_j = E_{\text{text}}(z_j;\phi)$.

\noindent\textbf{Step 2: Noise Generator}\\
Generate Gaussian parameters: $(\mu_j,\text{var}_j) = G_\theta(f_j,t_j)$.\\
Sample $\xi_j$ for $f_j$ from $\mathcal{N}(0,I_d)$.\\
Reparameterization: $\varepsilon_j=\mu_j+\text{var}_j\odot \xi_j$.

\noindent\textbf{Step 3: Compute Losses}\\
Compute base model classification loss $\mathcal{L}_\text{base} = \ell\big(h(f_j),\;y_j\big)$.\\
Compute variational proxy task loss $\mathcal{L}_\text{VPN} = \ell\big(h_{\text{noise}}(f_j+\varepsilon_j),\; y_j)$.

\noindent\textbf{Step 4: Update Parameters}\\
Update $\varphi$, $\theta$, $h$, and $h_{\text{noise}}$ by minimizing $\mathcal{L} = \mathcal{L}_\text{base} + \lambda\; \mathcal{L}_\text{VPN}$.

\noindent\textbf{Step 5: Repeat}\\
Repeat Steps 1–4 for each batch until convergence.
\end{algorithm}

\noindent\textbf{Loss Definitions.}
By Eqs.~(\ref{eq:task}) and (\ref{eq:vpn}), optimization is expressed as cross-entropy over the clean feature $f$ and its perturbed counterpart $\tilde f$. Unlike training with only perturbed features, we also incorporate clean features during detection training to mitigate the uncertainty introduced by the perturbations during inference. Thus, we use a clean head $h(\cdot)$ on $f$ and a noisy head $h_{\text{noise}}(\cdot)$ on $\tilde f=f+\varepsilon$. 

The training objective can be formulated as follows:
\begin{equation}
\mathcal{L}
=\underbrace{\ell\big(h(f),y\big)}_{\mathcal L_{\text{base}}}
+\lambda\ \underbrace{\ell\big(h_{\text{noise}}(f+\mu+\mathrm{var}\odot\xi),y\big)}_{\mathcal L_{\text{VPN}}},
\label{eq:final_loss}
\end{equation}
where $\ell$ means the BCELoss.

\noindent\textbf{The Parameters Update.}
We first define the feature-space gradients of the two branches:
\begin{gather}
g_{\text{clean}} := \frac{\partial\,\ell\big(h(f),y\big)}{\partial f}, \\
g_{\text{noisy}} := \frac{\partial\,\ell\big(h_{\text{noise}}(\tilde f),y\big)}{\partial \tilde f}, \quad \tilde f=f+\varepsilon. \nonumber
\end{gather}
Since $\varepsilon$ depends on $f$ via the noise generator $G_\theta(f, t)$, the chain rule yields the total gradient w.r.t.\ $f$:
\begin{equation}
\label{eq:gfinal-compact}
g_{\text{final}} \;=\; g_{\text{clean}} \;+\;
\lambda\,\!\left[
\underbrace{g_{\text{noisy}}}_{\partial \ell/\partial \tilde f}
\cdot
\underbrace{\Big(\frac{\partial \tilde f}{\partial f}\Big)}_{I\;+\;\frac{\partial \varepsilon}{\partial f}}
\right],
\end{equation}
where
\begin{equation}
\frac{\partial \varepsilon}{\partial f} \;=\; \frac{\partial \mu}{\partial f} \;+\; \operatorname{Diag}(\xi)\, \frac{\partial\,\mathrm{var}}{\partial f}.
\end{equation}

By the chain rule, the update of the PE's image encoder (LoRA) parameters $\varphi$ can be expressed as:
\begin{equation}
\varphi \;\leftarrow\; \varphi \;-\; \eta\,
\Big(\frac{\partial E_{\mathrm{img}}(x;\varphi )}{\partial \varphi}\Big)^{\!\top}\, g_{\mathrm{final}},
\end{equation}
where $\eta$ denotes the learning rate.

The dual constraints from $\mathcal L_{\text{base}}$ and $\mathcal L_{\text{VPN}}$ jointly shape the feature extractor: the clean term preserves performance on unperturbed features, while the noisy term provides a task-anchored, artifact-conditional signal that suppresses shortcut-sensitive directions and amplifies stable forensic cues, yielding consistent and trustworthy predictions. Finally, we provide an algorithm illustration of the proposed approach in Alg. \ref{alg:pin-clip-train} for an overall understanding.

During training, we update only the LoRA of the PE's image encoder ($\varphi$), the noise generator $\theta$ and the two linear classification heads $h$ and $h_{\text{noise}}$, keeping all other parameters frozen. At inference time, we discard the noise and use $E_{\mathrm{img}}$ with the clean head $h$ only.

\section{Experiments}
In this section, we provide an extensive evaluation of our method. We first introduce the experiment setup and then detail the results and observations.
\subsection{Experimental Setup}
\noindent\textbf{Training Datasets.}
Consistent with many methods \cite{chen2024drct,chen2025dual,guillaro2025bias}, we mitigate semantic shortcuts at the data level by employing VAE reconstruction to construct the training dataset. Specifically, we use MSCOCO \cite{lin2014microsoft} for real images and reconstruct them using VAE model in SDv2.1, ensuring that the reconstructed image size matches the real one.

\begin{table*}[!ht]
    \centering
    \Large
    \caption{\textbf{Cross-model Accuracy (Acc.) Performance on the AIGCDetect \cite{zhong2023patchcraft} Dataset.} Bold numbers indicate the best performance in each column, and underlined numbers indicate the second-best performance.}
    \label{table:GenImage_Acc}
    \resizebox{1.0\linewidth}{!}{
        \renewcommand{\arraystretch}{1.1}  
        \begin{tabular}{lcccccccccccccccccc}
        \toprule
        Method  &\rot{ProGAN} &\rot{Cyclegan}& \rot{BigGAN}& \rot{StyleGAN}&\rot{StyleGAN2}&\rot{GauGAN} &\rot{StarGAN} &\rot{WFIR} &\rot{SDv1.4} & \rot{{SDv1.5}} & \rot{{ADM}} & \rot{{GLIDE}}& \rot{{Midjourney}}& \rot{{Wukong}}& \rot{{VQDM}}& \rot{{DALLE2}} & \rot{\textit{mAcc.}} \\ \midrule
        CNN-Spot \cite{wang2020cnn}         & 49.9      & 49.5      & 49.8      & 49.9      & 49.9      & 50.0      & 50.0      & 50.0      & 93.9      & 94.1      & 51.0      & 54.6      & 97.7      & 91.2      & 51.5      & 52.0      & 61.6      \\
        UnivFD \cite{ojha2023towards}       & 86.5      & 97.0      & 82.2      & 84.6      & 75.2      & 84.0      & 95.9      & 98.9      & 93.9      & 93.8      & 80.9      & 66.4      & 89.7      & 92.8      & 92.1      & 93.9      & 88.0      \\ 
        NPR \cite{tan2024rethinking}        & 76.1      & 56.5      & 55.0      & 69.8      & 74.3      & 47.4      & 77.3      & 50.1      & 93.8      & 93.9      & 67.7      & \textbf{98.9}      & \underline{99.7}      & 83.0      & 65.8      & 97.3      & 75.4      \\
        DFFreq \cite{yan2026dual}  & 89.9      & 91.1      & 89.3      & \underline{91.1}      & \underline{90.0}      & 86.2      & 91.2      & 52.1      & 98.7      & 98.7      & 83.1      & 93.7      & \textbf{99.7}      & 96.1      & 93.1      & 98.6      & \underline{90.2}      \\
        AIDE \cite{yan2024sanity}           & 64.3      & 57.6      & 55.7      & 63.2      & 62.2      & 56.6      & 75.8      & 58.7      & 85.6      & 85.1      & 75.0      & 93.3      & 92.1      & 86.3      & 84.8      & 79.1      & 73.5      \\
        SAFE \cite{li2025improving}         & 72.6      & 69.9      & 65.9      & 77.4      & 67.2      & 55.8      & 79.8      & 50.5      & 95.9      & 96.0      & 56.3      & 91.7      & 98.9      & 89.7      & 88.6      & 91.1      & 78.0      \\
        VIB-Net \cite{zhang2025towards}     & 78.2      & 96.0      & 77.7      & 79.7      & 73.5      & 83.9      & 73.6      & 97.3      & 91.9      & 91.7      & 84.2      & 73.2      & 89.6      & 91.3      & 90.4      & 92.5      & 85.3      \\
        B-Free \cite{guillaro2025bias}      & \underline{95.5}      & 74.5      & \underline{91.5}      & 72.1      & 71.6      & \underline{96.6}      & 84.8      & 99.7      & 98.0      & 97.8      & 77.5      & 83.0      & 93.7      & 97.9      & 87.8      & 83.5      & 85.0      \\
        Effort \cite{yan2024orthogonal}     & 67.4      & \underline{98.4}      & 86.8      & 75.0      & 66.0      & 73.0      & \underline{97.2}      & \underline{100.0}     & \underline{99.0}      & \underline{98.8}      & \textbf{89.9}      & 92.4      & 98.8      & \underline{98.6}      & \underline{97.7}      & \textbf{99.9}      & 89.9      \\
        DDA \cite{chen2025dual}             & 84.7      & 61.4      & 76.2      & 72.7      & 81.4      & 88.1      & 61.6      & 51.6      & 98.0      & 98.1      & \underline{88.4}      & 86.5      & 96.3      & 97.7      & 63.1      & 94.2      & 81.3      \\ \midrule
        \rowcolor{light-gray} \textbf{PiND (Ours)}          & \textbf{95.6}      & \textbf{99.3}      & \textbf{99.6}      & \textbf{93.0}      & \textbf{93.4}      & \textbf{99.8}      & \textbf{99.8}      & \textbf{100.0}     & \textbf{99.9}      & \textbf{99.8}      & 84.6      & \underline{97.9}      & 98.0      & \textbf{99.7}      & \textbf{98.5}      & \underline{99.5}      & \textbf{97.4}      \\
        \bottomrule
        \end{tabular}
    }
\end{table*}

\begin{table*}[!ht]
    \centering
    \Large
    \caption{\textbf{Cross-model Average Precision (A.P.) Performance on the AIGCDetect \cite{zhong2023patchcraft} Dataset.} }
    \label{table:GenImage_AP}
    \resizebox{1.0\linewidth}{!}{
        \renewcommand{\arraystretch}{1.1}  
        \begin{tabular}{lcccccccccccccccccc}
        \toprule
        Method  &\rot{ProGAN} &\rot{Cyclegan}& \rot{BigGAN}& \rot{StyleGAN}&\rot{StyleGAN2}&\rot{GauGAN} &\rot{StarGAN} &\rot{WFIR} &\rot{SDv1.4} & \rot{{SDv1.5}} & \rot{{ADM}} & \rot{{GLIDE}}& \rot{{Midjourney}}& \rot{{Wukong}}& \rot{{VQDM}}& \rot{{DALLE2}} & \rot{\textit{mA.P.}} \\ \midrule
        CNN-Spot \cite{wang2020cnn}         & 42.9      & 38.6      & 40.9      & 34.3      & 38.8      & 36.6      & 45.9      & 32.5      & 99.7      & 99.7      & 67.2      & 86.8      & 99.9      & 99.4      & 71.2      & 91.6      & 64.1      \\
        UnivFD \cite{ojha2023towards}       & 95.8      & 99.7      & 97.1      & 90.5      & 81.1      & 97.5      & 99.4      & 99.9      & 99.2      & 98.9      & 89.6      & 77.2      & 95.6      & 98.4      & 98.0      & 98.3      & 94.8      \\
        NPR \cite{tan2024rethinking}        & 86.6      & 72.1      & 56.8      & 79.8      & 85.4      & 37.7      & 88.2      & 83.1      & 99.5      & 99.5      & 84.7      & \underline{99.9}      & 100.0     & 97.1      & 95.4      & 100.0     & 85.4      \\
        DFFreq \cite{yan2026dual}  & 98.0      & 99.5      & 97.3      & \underline{98.0}      & \underline{97.9}      & 94.7      & 99.7      & 73.3      & 99.9      & 99.9      & \textbf{99.0}      & 99.5      & \underline{100.0}     & 99.8      & \underline{99.8}      & 100.0     & \underline{97.3}      \\
        AIDE \cite{yan2024sanity}           & 71.5      & 67.4      & 59.2      & 69.4      & 69.5      & 60.3      & 87.8      & 65.4      & 95.9      & 95.5      & 89.4      & 98.3      & 99.0      & 96.0      & 94.7      & 95.0      & 82.1      \\
        SAFE \cite{li2025improving}         & 93.0      & 92.4      & 81.5      & 93.9      & 90.6      & 63.5      & 92.0      & 70.6      & 99.0      & 99.9      & 88.9      & 99.7      & \textbf{100.0}     & 99.4      & 99.6      & 99.9      & 91.6      \\
        VIB-Net \cite{zhang2025towards}     & 91.9      & 99.3      & 95.5      & 91.2      & 87.6      & 98.2      & 96.2      & 99.7      & 99.4      & 99.2      & 92.7      & 81.6      & 96.7      & 99.0      & 98.4      & 97.7      & 95.3      \\
        B-Free \cite{guillaro2025bias}      & \underline{98.6}      & 86.9      & \underline{97.7}      & 86.6      & 86.1      & \underline{99.2}      & 88.8      & 99.7      & 99.8      & 99.8      & 91.0      & 94.6      & 98.5      & 99.9      & 96.2      & 96.4      & 93.1      \\
        Effort \cite{yan2024orthogonal}     & 85.2      & \underline{99.8}      & 97.3      & 94.5      & 76.5      & 84.7      & \underline{100.0}     & \underline{100.0}     & 99.9      & \underline{100.0}     & 98.6      & 98.1      & 99.9      & 99.9      & 99.7      & \underline{100.0}     & 95.9      \\
        DDA \cite{chen2025dual}             & 97.7      & 72.3      & 88.0      & 84.2      & 91.8      & 95.2      & 66.7      & 68.8      & \underline{99.9}      & 99.9      & \underline{97.1}      & 96.8      & 99.6      & \underline{99.9}      & 81.3      & 98.8      & 89.9      \\ \midrule
        \rowcolor{light-gray} \textbf{PiND (Ours)}  & \textbf{99.9}      & \textbf{100.0}     & \textbf{100.0}     & \textbf{99.8}      & \textbf{99.4}      & \textbf{100.0}     & \textbf{100.0}     & \textbf{100.0}     & \textbf{100.0}     & \textbf{100.0}     & 97.7      & \textbf{99.9}      & 99.9      & \textbf{100.0}     & \textbf{99.9}      & \textbf{100.0}     & \textbf{99.8}      \\
        \bottomrule
        \end{tabular}
    }
\end{table*}

\noindent\textbf{Evaluation Datasets.}
\textbf{I) Ideal benchmark.} We first evaluate PiND on two clean and high-quality datasets, AIGCDetect \cite{zhong2023patchcraft} and AIGIBench \cite{li2025artificial}. AIGCDetect \cite{zhong2023patchcraft} contains 15 subsets derived from different kinds of generative models, including ProGAN \cite{karras2018progressive}, CycleGAN \cite{zhu2017unpaired}, BigGAN \cite{brock2018large}, StyleGAN \cite{karras2019style}, StyleGAN2 \cite{karras2020analyzing}, GauGAN \cite{park2019semantic}, StarGAN \cite{choi2018stargan}, WFIR \cite{WFIR}, SDv1.4 \cite{rombach2022high}, SDv1.5 \cite{rombach2022high}, ADM \cite{dhariwal2021diffusion}, GLIDE \cite{nichol2021glide}, Midjourney \cite{midjourney}, Wukong \cite{wukong}, VQDM \cite{gu2022vector} and DALLE2 \cite{ramesh2022hierarchical}, and Deepfake \cite{rossler2019faceforensics++}. AIGIBench comprehensively simulates state-of-the-art image generation methods, including ProGAN \cite{karras2018progressive}, R3GAN \cite{huang2024gan}, StyleGAN3 \cite{karras2021alias}, StyleGAN-XL \cite{sauer2022stylegan}, StyleSwim \cite{zhang2022styleswin}, WFIR \cite{WFIR}, BlendFace\cite{shiohara2023blendface}, E4S \cite{liu2023fine}, FaceSwap \cite{Faceswap}, InSwap \cite{inswapper}, SimSwap \cite{chen2020simswap}, DALLE-3 \cite{dalle3}, FLUX1-dev \cite{FLUX_1}, Midjourney-V6 \cite{midjourney}, GLIDE \cite{nichol2021glide}, Imagen3 \cite{Imagen_3}, SD3 \cite{esser2024scaling}, SDXL \cite{podell2023sdxl}, BLIP \cite{li2022blip}, Infinite-ID \cite{wu2024infinite}, PhotoMaker \cite{li2024photomaker}, Instant-ID \cite{wang2024instantid}, IP-Adapter \cite{ye2023ip}, CommunityAI and SocialRF.
\textbf{II) Real-world degradation benchmark.} We also evaluate PiND on both public and newly collected real-world datasets, including Chameleon \cite{yan2024sanity}, SynthWildx \cite{cozzolino2024raising}, WildRF \cite{cavia2024real}. These datasets include various propagation environments, black-box post-processing methods, and image formats produced by both open-source and commercial models. Together, they represent the complexities of real-world social media situations and are used to assess the robustness and generalization capabilities of models under uncontrolled degradation conditions.

\begin{table*}[ht!]
    \caption{\textbf{Cross-model Accuracy (Acc.) \& Average Precision (A.P.) Performance on the AIGIBench Datasets \protect \cite{li2025artificial}.}}
    \label{table_AIGIBench}
    \large
    \renewcommand\arraystretch{1.1}
    \resizebox{1.0\linewidth}{!}{
        \begin{tabular}{lcccccccccccccccccc}
        \toprule
        \multirow{2}{*}{Method} & \multicolumn{2}{c}{ProGAN} & \multicolumn{2}{c}{R3GAN}  & \multicolumn{2}{c}{StyleGAN3} & \multicolumn{2}{c}{StyleGAN-XL} & \multicolumn{2}{c}{StyleSwim} & \multicolumn{2}{c}{WFIR}     & \multicolumn{2}{c}{BlendFace} & \multicolumn{2}{c}{E4S} & \multicolumn{2}{c}{FaceSwap}  \\
        \cmidrule{2-19}
                                & Acc.         & A.P.        & Acc.         & A.P.        & Acc.          & A.P.          & Acc.         & A.P.             & Acc.          & A.P.          & Acc.          & A.P.         & Acc.         & A.P.            & Acc.         & A.P.    & Acc.         & A.P.           \\ \midrule
        CNN-Spot                & 49.9         & 42.9        & 49.7         & 43.5        & 50.2          & 56.4          & 49.8         & 36.5             & 50.8          & 66.9          & 50.0          & 32.5         & 50.3         & 75.9            & 49.9         & 65.4    & 50.6         & 60.1           \\
        UnivFD                  & 86.5         & 95.8        & 62.7         & 72.9        & 63.5          & 68.6          & 65.7         & 65.9             & 66.4          & 79.7          & 98.7          & 99.9         & 59.7         & 69.1            & 64.9         & 80.5    & 66.1         & 82.0           \\
        NPR                     & 76.1         & 86.1        & 53.6         & 65.7        & 64.6          & 79.5          & 89.6         & \underline{96.0}             & 85.2          & \underline{95.0}          & 50.1          & 83.1         & 53.8         & 57.7            & 50.1         & 54.8    & 72.9         & \textbf{90.0}           \\
        DFFreq                  & \underline{89.9}         & \underline{98.0}        & 87.0         & 85.2        & \underline{89.6}          & \underline{94.5}          & 64.4         & 68.4             & \underline{89.8}          & 89.0          & 52.1          & 73.3         & 33.5         & 42.9            & 34.3         & 43.2    & 62.8         & 71.3           \\
        AIDE                    & 64.3         & 71.5        & 65.0         & 75.7        & 61.4          & 72.3          & 54.7         & 56.9             & 74.1          & 83.9          & 58.7          & 65.4         & \underline{71.6}         & \underline{82.9}            & 67.5         & 81.0    & 71.4         & \underline{85.4}           \\
        SAFE                    & 72.6         & 93.0        & 84.1         & 92.9        & 69.4          & 79.8          & \underline{94.5}         & 89.7             & 86.8          & 85.6          & 50.5          & 70.6         & 43.6         & 46.7            & 43.9         & 47.5    & 58.3         & 76.2           \\
        VIB-Net                 & 78.1         & 91.9        & 64.0         & 77.0        & 66.8          & 69.4          & 58.9         & 58.5             & 64.6          & 66.3          & 97.3          & 99.7         & 65.1         & 71.5            & 68.0         & 80.8    & 67.7         & 82.2           \\
        Effort                  & 67.4         & 85.2        & 80.9         & 87.8        & 83.5          & 85.2          & 62.1         & 55.4             & 79.6          & 78.3          & \underline{100.0}         & \underline{100.0}        & 72.0         & 65.5            & 69.5         & 58.9    & \textbf{82.3}         & 72.5           \\
        DDA                     & 84.7         & 97.7        & \underline{93.9}         & \underline{98.6}        & 58.3          & 76.2          & 46.6         & 35.9             & 56.6          & 68.5          & 51.6          & 68.8         & \textbf{77.9}         & \textbf{90.0}            & \underline{82.3}         & \underline{87.8}    & \underline{77.3}         & 84.1           \\ \midrule
    \rowcolor{light-gray}\textbf{PiND (Ours)}   & \textbf{95.6}         & \textbf{99.9}        & \textbf{95.5}         & \textbf{99.5}        & \textbf{98.6}          & \textbf{99.7}          & \textbf{95.4}         & \textbf{97.9}             & \textbf{96.3}          & \textbf{99.9}          & \textbf{100.0}         & \textbf{100.0}        & 61.2         & 84.9            & \textbf{91.8}         & \textbf{95.4}    & 69.4         & 83.6           \\ \midrule
        \end{tabular}
    }
    \renewcommand\arraystretch{1.2}
    \resizebox{1.0\linewidth}{!}{
        \begin{tabular}{lcccccccccccccccccc}
        \hline
        \multirow{2}{*}{Method} & \multicolumn{2}{c}{InSwap} & \multicolumn{2}{c}{SimSwap}   & \multicolumn{2}{c}{FLUX1-dev}   & \multicolumn{2}{c}{Midjourney V6}  & \multicolumn{2}{c}{GLIDE}    & \multicolumn{2}{c}{DALLE-3}    & \multicolumn{2}{c}{Imagen3} & \multicolumn{2}{c}{SD3}    & \multicolumn{2}{c}{SDXL}\\
         \cline{2-19} 
                                & Acc.         & A.P.        & Acc.          & A.P.          & Acc.         & A.P.             & Acc.          & A.P.               & Acc.          & A.P.         & Acc.         & A.P.            & Acc.         & A.P.         & Acc.         & A.P.         & Acc.         & A.P.     \\ \midrule
        CNN-Spot                & 50.7         & 71.4        & 50.1          & 66.7          & 61.8         & 91.0             & 84.9          & 94.7               & 53.7          & 76.4         & 76.6         & 96.7            & 78.5         & 98.5         & 73.7         & 96.9         & \underline{96.5}         & 99.8     \\
        UnivFD                  & 60.0         & 67.5        & 65.5          & 77.1          & 46.9         & 44.2             & 54.9          & 67.4               & 40.5          & 41.5         & 52.1         & 63.1            & 64.4         & 77.0         & 68,3         & 75.6         & 70.8         & 84.5     \\
        NPR                     & 67.3         & 82.7        & 60.2          & 76.5          & 93.0         & 99.0             & 82.2          & 96.5               & \textbf{95.2}          & \textbf{99.5}         & 45.5         & 53.6            & \underline{93.9}         & \underline{99.0}         & 81.2         & 93.4         & 95.7         & 99.9     \\
        DFFreq                  & 45.4         & 61.5        & 44.7          & 61.6          & 89.5         & 96.3             & 71.0          & 91.6               & 85.8          & 92.2         & 41.6         & 51.4            & 84.8         & 86.7         & 91.4         & 95.5         & 92.2         & 98.4     \\
        AIDE                    & 70.2         & 84.0        & 71.7          & 85.0          & 78.6         & 88.5             & 88.3          & 94.9               & 90.7          & 97.2         & 56.3         & 65.9            & 81.0         & 91.0         & 61.2         & 67.8         & 77.1         & 88.1     \\
        SAFE                    & 51.7         & 69.9        & 48.3          & 64.3          & 93.0         & 97.4             & 84.9          & 97.3               & 87.1          & 92.8         & 45.8         & 51.7            & 92.4         & 98.1         & 84.7         & 93.1         & 95.8         & 99.7     \\
        VIB-Net                 & 64.2         & 72.2        & 66.7          & 76.7          & 55.6         & 64.3             & 63.5          & 84.1               & 49.8          & 52.6         & 57.2         & 81.7            & 66.7         & 87.7         & 68.9         & 84.5         & 71.7         & 94.6     \\
        Effort                  & \underline{77.5}         & 69.7        & \underline{82.3}          & 75.3          & 80.8         & 94.8             & 68.5          & 96.2               & 76.7          & 73.9         & 67.6         & 73.2            & 82.3         & 98.9         & 81.7         & 97.1         & 82.8         & 99.0     \\
        DDA                     & \textbf{82.0}         & \textbf{87.8}        & \textbf{83.3}          & \textbf{89.3}          & \underline{95.2}         & \underline{99.3}             & \underline{96.4}          & \underline{99.5}               & 85.1          & 93.7         & \underline{93.1}         & \underline{98.1}            & 90.9         & 97.2         & \underline{95.7}         & \underline{100.0}        & 95.8         & \underline{100.0}    \\ \midrule
    \rowcolor{light-gray}\textbf{PiND (Ours)}   & 70.7         & \underline{87.3}        & 73.1          & \underline{87.7}          & \textbf{95.3}         & \textbf{99.9}             & \textbf{97.4}          & \textbf{99.8}               & \underline{93.9}          & \underline{98.7}         & \textbf{95.6}         & \textbf{99.6}            & \textbf{95.9}         & \textbf{99.8}         & \textbf{96.7}         & \textbf{100.0}        & \textbf{96.7}         & \textbf{100.0}    \\ \midrule
        \end{tabular}
    }
    \renewcommand\arraystretch{1.2}
    \resizebox{1.0\linewidth}{!}{
        \begin{tabular}{lcccccccccccccc|cc}
        \hline
        \multirow{2}{*}{Method} & \multicolumn{2}{c}{BLIP}      & \multicolumn{2}{c}{Infinite-ID}   & \multicolumn{2}{c}{InstantID} & \multicolumn{2}{c}{IP-Adapter}  & \multicolumn{2}{c}{PhotoMaker}     & \multicolumn{2}{c}{SocialRF} & \multicolumn{2}{c}{CommunityAI}     & \multicolumn{2}{|c}{Mean}    \\
         \cline{2-17} 
                                & Acc.         & A.P.           & Acc.         & A.P.               & Acc.          & A.P.          & Acc.         & A.P.             & Acc.          & A.P.               & Acc.          & A.P.         & Acc.         & A.P.                 & Acc.         & A.P.         \\ \midrule
        CNN-Spot                & \textbf{99.3}         & 100.0          & 72.9         & 96.6               & 95.2          & 99.9          & 83.8         & 98.8             & 50.6          & 72.5               & 65.4          & 79.6         & 62.3         & 89.1                 & 64.3         & 76.4         \\
        UnivFD                  & 68.5         & 89.3           & 69.4         & 92.0               & 81.7          & 94.4          & 66.7         & 79.7             & 66.8          & 85.2               & 53.9          & 57.6         & 60.0         & 66.0                 & 64.4         & 74.6         \\
        NPR                     & 96.3         & 99.9           & 94.0         & 99.5               & 94.4          & 99.9          & 94.5         & \underline{99.8}             & 94.3          & 99.7               & 56.6          & 65.1         & 54.8         & 58.2                 & 75.8         & 85.2         \\
        DFFreq                  & \underline{99.0}         & \underline{100.0}          & 91.0         & 96.1               & 89.8          & 98.1          & 89.8         & 96.3             & 89.7          & 97.6               & 59.3          & 62.8         & 55.0         & 58.1                 & 72.9         & 80.4         \\
        AIDE                    & 90.1         & 96.8           & 76.4         & 87.8               & 83.1          & 93.1          & 76.8         & 88.0             & 67.9          & 80.5               & 56.5          & 62.3         & 63.7         & 75.2                 & 71.1         & 80.8         \\
        SAFE                    & 98.9         & 99.7           & 95.1         & 99.8               & 94.8          & 98.6          & 93.9         & 98.4             & \underline{94.6}          & \underline{99.8}               & 57.5          & 64.6         & 54.5         & 47.4                 & 75.1         & 82.2         \\
        VIB-Net                 & 66.5         & 91.7           & 69.1         & 89.5               & 67.9          & 86.4          & 67.2         & 87.3             & 67.3          & 85.1               & 56.9          & 63.4         & 56.0         & 61.1                 & 65.8         & 78.4         \\
        Effort                  & 90.0         & 98.7           & 82.5         & 99.3               & 82.3          & 97.8          & 82.3         & 97.9             & 82.1          & 96.5               & 55.4          & 57.9         & 47.4         & 44.7                 & 76.8         & 82.4         \\
        DDA                     & 95.9         & 99.9           & \underline{96.5}         & \underline{100.0}              & \textbf{96.7}          & \underline{99.9}          & \underline{95.0}         & 99.1             & 78.7          & 90.8               & \underline{79.7}          & \underline{88.9}         & \underline{88.5}         & \underline{95.7}                 & \underline{83.1}         & \underline{89.9}         \\ \midrule
    \rowcolor{light-gray}\textbf{PiND (Ours)}   & 95.6         & \textbf{100.0}          & \textbf{96.5}         & \textbf{100.0}              & \underline{95.9}          & \textbf{100.0}         & \textbf{96.0}         & \textbf{100.0}            & \textbf{95.8}          & \textbf{99.9}               & \textbf{92.2}          & \textbf{95.0}         & \textbf{89.5}         & \textbf{98.0}                 & \textbf{91.2}         & \textbf{97.1}         \\ \bottomrule
        \end{tabular}
    }
\end{table*}

\noindent\textbf{Baseline Detectors.} We evaluate 10 off-the-shelf detectors including CNN-Spot \cite{wang2020cnn} (CVPR 2020), UnivFD \cite{ojha2023towards} (CVPR 2023), NPR \cite{tan2024rethinking} (CVPR 2024), AIDE \cite{yan2024sanity} (ICLR 2025), DFFreq \cite{yan2026dual} (TIFS 2026), SAFE \cite{li2025improving} (KDD 2025), VIB-Net \cite{zhang2025towards} (CVPR 2025), B-Free \cite{guillaro2025bias} (CVPR 2025), Effort \cite{yan2024orthogonal} (ICML 2025), and DDA \cite{chen2025dual} (NeurIPS 2025) for comparison. To ensure a fair comparison, we retrained all methods using the aforementioned training set, except for B-Free and DDA, whose results are obtained directly from the official implementations and pretrained weights released by their authors.

\noindent\textbf{Implementation Details.}
we adopt PE-Core-L14-336 \cite{bolya2025perception} as the backbone network and finetune it using the LoRA \cite{hu2022lora} algorithm, using the rank of 8. During training, we use randomly cropped windows of $336 \times 336$, while we use center-cropped windows of $336 \times 336$ during inference. Padding is applied when the image height or width is insufficient. The hyperparameter $\lambda$ is set to 0.2. No data augmentation is applied in our experiments. During training, we use the Adam optimizer \cite{kingma2014adam}, the batch size is set to 64, and the training lasts for only 1 epoch. Our proposed method is implemented using the PyTorch \cite{paszke2019pytorch} library. All experiments are conducted on an NVIDIA RTX 4090 GPU (48G).

\begin{table*}[t]
    \centering
    \caption{\textbf{Cross-model Accuracy (Acc.) on Real-world Datasets.} Datasets include Chameleon \cite{yan2024sanity}, SynthWildX \cite{cozzolino2024raising} and WildRF \cite{cavia2024real}.}
    \label{tab:real-world}
    \renewcommand{\arraystretch}{1.1}  
    \resizebox{1.0\linewidth}{!}{
        \begin{tabular}{l|c|ccc|ccc|cc}
        \midrule
        \multirow{2}{*}{Method}     & \multirow{2}{*}{Chameleon}    & \multicolumn{3}{c|}{SynthWildX}       & \multicolumn{3}{c|}{WildRF}             & \multicolumn{2}{c}{Mean}    \\ \cmidrule{3-10}
                                    & ~                             & DALLE-3   & Firefly   & Midjourney    & Facebook      & Reddit    & Twitter     & Acc.        & A.P.          \\ \midrule
        CNN-Spot \cite{wang2020cnn} & 65.7                          & 82.2      & 50.1      & 88.6          & 84.1          & 67.5      & 68.4        & 72.3        & 84.2          \\
        UnivFD \cite{ojha2023towards}& 62.8                          & 55.3      & 56.6      & 47.5          & 47.2          & 66.0      & 57.1        & 56.1        & 58.6          \\
        NPR \cite{tan2024rethinking}& 59.4                          & 49.8      & 50.0      & 50.2          & 50.3          & 67.3      & 50.3        & 53.9        & 56.4          \\
        DFFreq \cite{yan2026dual}& 59.4                          & 56.0      & 56.6      & 56.0          & 62.9          & 75.3      & 40.9        & 58.2        & 51.4          \\
        AIDE \cite{yan2024sanity}   & 62.6                          & 55.2      & 49.4      & 78.4          & 60.3          & 61.7      & 63.3        & 61.6        & 67.2          \\
        SAFE \cite{li2025improving} & 59.1                          & 49.2      & 50.1      & 49.3          & 50.9          & 70.5      & 38.5        & 52.5        & 58.7          \\
        VIB-Net \cite{zhang2025towards}& 60.8                          & 63.3      & 57.1      & 51.1          & 52.5          & 68.9      & 56.2        & 58.6        & 66.0          \\
        B-Free \cite{guillaro2025bias}& 78.0                          & 88.4      & 82.9      & 87.6          & \underline{87.8}          & 76.5      & 88.6        & 84.3        & 93.3          \\
        Effort \cite{yan2024orthogonal}& 57.0                          & 52.1      & 55.1      & 47.0          & 60.3          & 71.5      & 47.0        & 55.7        & 62.5          \\
        DDA \cite{chen2025dual}     & \underline{84.8}                          & \underline{91.0}      & \underline{84.6}      & \underline{91.8}          & 86.9          & \underline{82.4}      & \underline{88.9}        & \underline{87.2}        & \underline{94.6}          \\ \midrule
        \rowcolor{light-gray}\textbf{PiND (Ours)}   & \textbf{92.4}                          & \textbf{97.0}      & \textbf{93.3}      & \textbf{95.5}          & \textbf{96.9}          & \textbf{97.9}      & \textbf{97.7}        & \textbf{95.8}        & \textbf{98.5}          \\ \midrule
        \end{tabular}
    }
\end{table*}

\begin{table*}[t]
\centering
\Large
\caption{\textbf{Ablation Studies on the Noise Generator.}}
\label{tab:Ablation}
\resizebox{1.0\linewidth}{!}{
    \renewcommand{\arraystretch}{1.1}  
    \begin{tabular}{c|ccccc|cc}
    \toprule
    \textit{Noise Generator}                 & \textit{AIGCDetect} \cite{zhong2023patchcraft}    & \textit{AIGIBench} \cite{li2025artificial}    & \textit{Chameleon} \cite{yan2024sanity}    & \textit{SynthWildX} \cite{cozzolino2024raising}   & \textit{WildRF} \cite{cavia2024real}       & \textit{Mean Acc.}     & \textit{Mean A.P.} \\ \midrule
    \xmark                          & 95.6          & 85.7          & 77.5          & 91.0          & 91.6          & 88.3      & 97.0      \\ \midrule
    Random                          & 96.4          & 87.5          & 83.7          & 92.1          & 91.6          & 90.3      & 97.1      \\
    Sample                          & 95.9          & 89.2          & 81.3          & 92.7          & 92.4          & 90.7      & 98.0      \\ \midrule
    MLP                             & 96.7          & 88.1          & \underline{90.9}          & \underline{93.7}          & 94.4          & \underline{92.8}      & 97.8      \\
    self-attention of text          & 97.0          & 89.8          & 87.6          & 92.5          & \underline{96.5}          & 92.7      & \underline{98.3}      \\
    self-attention of image         & \underline{97.3}          & \underline{90.8}          & 85.4          & 92.8          & 95.3          & 92.3      & 98.1      \\ \midrule
    \rowcolor{light-gray} \textbf{PiND (Ours)}      & \textbf{97.4}          & \textbf{91.2}          & \textbf{92.4}          & \textbf{95.3}          & \textbf{97.5}          & \textbf{94.8}      & \textbf{98.4}      \\ \bottomrule
    \end{tabular}
}
\end{table*}

\subsection{Effectiveness Evaluation}
\noindent\textbf{Performance on Ideal Benchmark.} The results in Table \ref{table:GenImage_Acc} and Table \ref{table:GenImage_AP} demonstrate the capability of our method on AIGCDetect \cite{zhong2023patchcraft}, with the highest Acc. 97.4\% and A.P. 99.8\% on average. Compared with the latest state-of-the-art method, DFFreq, our approach improves mAcc. by 7.2\% and mA.P. by 2.5\%. In addition, on the more comprehensive dataset AIGIBench, our method achieved an mAcc. of 85.8\% and mA.P. of 92.5\%, outperforming all competing methods by a notable margin, as shown in Table \ref{table_AIGIBench}. Compared with DDA, the strongest baseline, our method yields an additional 8.1\% gain in accuracy and 7.2\% improvement in A.P. While conventional detectors such as CNN-Spot and NPR perform well on specific generators (e.g., SDXL, BLIP), they fail to generalize to newer generation techniques, especially real-world datasets such as SocialRF and CommunityAI. In contrast, our method consistently maintains high accuracy across most models, indicating that task-beneficial perturbations effectively enhance the model's detection capability.

\noindent\textbf{Performance on Real-world Degradation Benchmark.} To further assess real-world effectiveness, we evaluated our method on the latest publicly available in-the-wild benchmarks, Chameleon \cite{yan2024sanity}, SynthWildX \cite{cozzolino2024raising} and WildRF \cite{cavia2024real}, which incorporate diverse compression settings and unknown post-processing operations. The results is shown in Table \ref{tab:real-world}. Across all three datasets, our method achieves an  mAcc. of 95.8\% and mA.P. of 98.5\%, surpassing the latest state-of-the-art method, DDA, by 8.6\% in mAcc. These results demonstrate that our method effectively accommodates common degradations and distribution shifts encountered in real-world network. Compared with most other methods that achieve around 60\% accuracy, both B-Free and DDA demonstrate strong performance, reaching over 80\% accuracy. This improvement is partly due to their alignment training strategy, which helps mitigate shortcuts. However, our method outperforms these approaches, achieving even higher accuracy, further demonstrating that our Positive-incentive Noise more effectively suppresses shortcut-sensitive directions while amplifying stable, task-relevant forensic evidence.

\begin{table*}[t]
\centering
\large
\caption{\textbf{Backbone Comparison across Five Datasets.} Our method effectively improves the detection performance of all backbones.}
\label{tab:backbone}
\resizebox{1.0\linewidth}{!}{
    \renewcommand{\arraystretch}{1.1}  
    \begin{tabular}{c|ccccc|cc}
    \toprule
    \textit{Backbone}                   & \textit{AIGCDetect} \cite{zhong2023patchcraft}    & \textit{AIGIBench} \cite{li2025artificial}    & \textit{Chameleon} \cite{yan2024sanity}    & \textit{SynthWildX} \cite{cozzolino2024raising}   & \textit{WildRF} \cite{cavia2024real}       & \textit{Mean Acc.}     & \textit{Mean A.P.} \\ \midrule
    CLIP:ViT-L/14                        & 88.9          & 74.3          & \textbf{59.7}          & 50.1          & 60.9          & 66.8      & 69.8  \\
    \rowcolor{light-gray}\textbf{+ Ours}         & \textbf{92.4}          & \textbf{79.2}          & 58.9          & \textbf{53.4}          & \textbf{64.5}          & \pmb{$69.7_{+2.9}$}      & \pmb{$70.7_{+0.9}$}  \\ \midrule
    PE-Core-B16-224                     & 82.9          & 65.9          & 62.8          & 50.3          & 57.1          & 63.8      & 69.5  \\
    \rowcolor{light-gray}\textbf{+ Ours}         & \textbf{91.9}          & \textbf{74.0}          & \textbf{65.7}          & \textbf{72.0}          & \textbf{76.1}          & \pmb{$75.9_{+12.1}$}      & \pmb{$82.5_{+13.0}$}  \\ \midrule
    PE-Core-L14-336                     & 95.6          & 85.7          & 77.5          & 91.0          & 91.6          & 88.3      & 97.0  \\
    \rowcolor{light-gray}\textbf{+ Ours}         & \textbf{97.4}          & \textbf{91.2}          & \textbf{92.4}          & \textbf{95.3}          & \textbf{97.5}          & \pmb{$94.8_{+6.5}$}      & \pmb{$98.4_{+4.8}$}  \\ \midrule
    PE-Core-G14-448                     & 95.6          & 85.5          & 69.4          & 81.7          & 83.6          & 83.2      & 92.9  \\
    \rowcolor{light-gray}\textbf{+ Ours}         & \textbf{98.5}          & \textbf{88.8}          & \textbf{70.6}          & \textbf{83.9}          & \textbf{89.4}          & \pmb{$86.2_{+3.0}$}      & \pmb{$97.7_{+4.8}$}  \\ \midrule
    \end{tabular}
}
\end{table*}

\begin{table*}[t]
\centering
\large
\caption{\textbf{Hyperparameter $\lambda$ Ablation Experiment on the Five Datasets.}}
\label{tab:lambda}
\resizebox{1.0\linewidth}{!}{
    \renewcommand{\arraystretch}{1.1}  
    \begin{tabular}{c|ccccc|cc}
    \toprule
    Hyperparameter $\lambda$        & \textit{AIGCDetect} \cite{zhong2023patchcraft}    & \textit{AIGIBench} \cite{li2025artificial}    & \textit{Chameleon} \cite{yan2024sanity}    & \textit{SynthWildX} \cite{cozzolino2024raising}   & \textit{WildRF} \cite{cavia2024real}       & \textit{Mean Acc.}     & \textit{Mean A.P.} \\ \midrule
    0.0                             & 95.6          & 85.7          & 77.5          & 91.0          & 91.6          & 88.3      & 97.0      \\
    0.1                             & \underline{97.3}          & 89.9          & \underline{89.9}          & \underline{94.3}          & \underline{96.5}          & \underline{93.6}      & \textbf{97.9}      \\ \midrule
    \rowcolor{light-gray} 0.2       & \textbf{97.4}          & \textbf{91.2}          & \textbf{92.4}          & \textbf{95.3}          & \textbf{97.5}          & \textbf{94.8}      & \underline{98.4}      \\ \midrule
    0.3                             & 96.1          & 90.5          & 83.9          & 90.9          & 91.9          & 90.7      & 97.6      \\
    0.4                             & 96.4          & 90.7          & 81.7          & 89.3          & 88.9          & 89.4      & 97.4      \\
    0.6                             & 96.3          & 91.0          & 78.0          & 86.0          & 87.1          & 87.7      & 96.6      \\
    0.8                             & 95.6          & \underline{91.2}          & 77.6          & 84.1          & 85.2          & 86.7      & 96.6      \\
    1.0                             & 96.4          & 90.9          & 74.3          & 80.6          & 83.2          & 85.1      & 96.2      \\ \bottomrule
    \end{tabular}
}
\end{table*}

\begin{table}[t]
    \centering
    \Large
    \caption{\textbf{Robustness on JPEG Compression and Gaussian Blur.} The accuracy (\%) averaged on AIGCDetect \cite{zhong2023patchcraft}.}
    \label{table:Robustness_GenImage}
    \resizebox{1.0\linewidth}{!}{
    \renewcommand{\arraystretch}{1.2}  
    \begin{tabular}{l|ccc|ccc}
    \toprule
    \multirow{2}{*}{\textbf{Method}}         & \multicolumn{3}{c|}{\textbf{JPEG Compression}}  & \multicolumn{3}{c}{\textbf{Gaussian Blur}}                     \\
                                            & \textbf{QF=95}         & \textbf{QF=75}             & \textbf{QF=60}         & \pmb{$\sigma = 0.5$}        & \pmb{$\sigma = 1.5$}        & \pmb{$\sigma = 2.5$}        \\ \midrule
    UnivFD  \cite{ojha2023towards}          & 73.9          & 69.6              & 67.9          & 78.2                  & 66.8                  & 64.7                  \\
    NPR \cite{tan2024rethinking}            & 56.7          & 59.0              & 58.6          & 72.7                  & 72.0                  & 72.5                  \\
    DFFreq \cite{yan2026dual}      & 64.1          & 63.0              & 62.1          & 90.1                  & 85.7                  & 81.8                  \\
    AIDE \cite{yan2024sanity}               & 56.6          & 58.4              & 57.3          & 67.3                  & 65.6                  & 67.4                  \\
    VIB-Net \cite{zhang2025towards}         & 70.8          & 67.9              & 63.0          & 78.0                  & 68.7                  & 68.7                  \\
    Effort \cite{yan2024orthogonal}         & \underline{81.8}          & 78.7              & 74.6          & \underline{91.3}                  & \underline{87.8}                  & \underline{87.1}                  \\
    DDA \cite{chen2025dual}                 & 80.7          & \underline{80.6}              & \underline{79.5}          & 80.6                  & 80.0                  & 80.3                  \\ \midrule
    \textbf{PiND (Ours)}                           & \textbf{95.2}          & \textbf{90.8} 	            & \textbf{88.7}          & \textbf{96.8}                  & \textbf{94.1}                  & \textbf{91.4}                  \\ \bottomrule
    \end{tabular}
    }
\end{table}

\begin{table}[t]
    \centering
    \Large
    \caption{\textbf{Robustness on JPEG Compression and Gaussian Blur.} The accuracy (\%) averaged on real-world datasets, including Chameleon \cite{yan2024sanity}, SynthWildX \cite{cozzolino2024raising} and WildRF \cite{cavia2024real}.}
    \label{table:Robustness_AIGIBench}
    \resizebox{1.0\linewidth}{!}{
    \renewcommand{\arraystretch}{1.2}  
    \begin{tabular}{l|ccc|ccc}
    \toprule
    \multirow{2}{*}{\textbf{Method}}         & \multicolumn{3}{c|}{\textbf{JPEG Compression}}       & \multicolumn{3}{c}{\textbf{Gaussian Blur}}                     \\
                                            & \textbf{QF=95}         & \textbf{QF=75}             & \textbf{QF=60}         & \pmb{$\sigma = 0.5$}        & \pmb{$\sigma = 1.5$}        & \pmb{$\sigma = 2.5$}        \\ \midrule
    UnivFD \cite{ojha2023towards}           & 50.0          & 50.5              & 51.0          & 54.7                  & 52.9                  & 53.3                  \\
    NPR \cite{tan2024rethinking}            & 62.2          & 60.1              & 58.1          & 57.1                  & 57.1                  & 57.4                  \\
    DFFreq \cite{yan2026dual}      & 60.6          & 57.9              & 56.3          & 58.5                  & 62.7                  & 62.9                  \\
    AIDE \cite{yan2024sanity}               & 60.0          & 60.0              & 59.9          & 61.2                  & 57.1                  & 58.5                  \\
    VIB-Net \cite{zhang2025towards}         & 62.2          & 60.5              & 58.2          & 64.2                  & 56.7                  & 56.5                  \\
    Effort \cite{yan2024orthogonal}         & 68.0          & 68.6              & 65.3          & 73.1                  & 73.8                  & 73.2                  \\
    DDA \cite{chen2025dual}                 & \underline{86.9}          & \underline{85.1}              & \underline{81.6}          & \underline{86.7}                  & \underline{87.3}                  & \underline{87.2}                  \\ \midrule
    \textbf{PiND (Ours)}                           & \textbf{94.8}          & \textbf{93.1}	            & \textbf{93.7}          & \textbf{95.9}                  & \underline{94.6}                  & \textbf{93.4}                  \\ \bottomrule
    \end{tabular}
    }
\end{table}

\subsection{Ablation Studies}
\label{sec:ablation}

\noindent\textbf{Ablation Studies of the Noise Generator.} We conduct additional analyzes on the Noise Generator, as summarized in Table~\ref{tab:Ablation}. Without any noise, the average accuracies on five datasets are 88.3\% and 97\%, respectively. Introducing our Positive-incentive Noise improves the results by 6.5\% and 1,4\%, confirming that task-beneficial perturbations effectively enhance PE’s detection capability. 

As discussed in Section \ref{sec:intro}, we argue that subtle feature-space perturbations can steer the model away from spurious shortcuts and toward causal, generalizable cues. To further support this claim, we evaluate two alternative perturbations: (i) i.i.d. random Gaussian noise and (ii) mean-aligned Gaussian noise (i.e., Gaussian noise shifted toward the visual feature mean). Both variants yield moderate improvements (and can sometimes surpass existing baselines), improving mean accuracy by 2.0\% and 2.4\% over the baseline, respectively. These results support our hypothesis that small feature-space perturbations can disproportionately disrupt brittle shortcut evidence. Nevertheless, compared with our PiND, these variants show noticeable performance degradation, further validating that our designed perturbation better suppresses shortcut-sensitive directions while enhancing stable forensic cues through beneficial stochastic transformations, resulting in more consistent and reliable predictions.

In addition, to assess the contribution of cross-attention fusion in the noise generator, we conducted additional ablation experiments. An MLP-based or self-attention-based generator conditioned only on visual features achieves competitive performance, with mean accuracies of 92.8\% and 92.3\%, respectively. However, when incorporating text features as label-level guidance, the generator can adapt its perturbations more precisely to the underlying forensic categories, yielding the best performance of approximately 94.8\%.

\begin{figure*}
    \centering
    \includegraphics[width=1\linewidth]{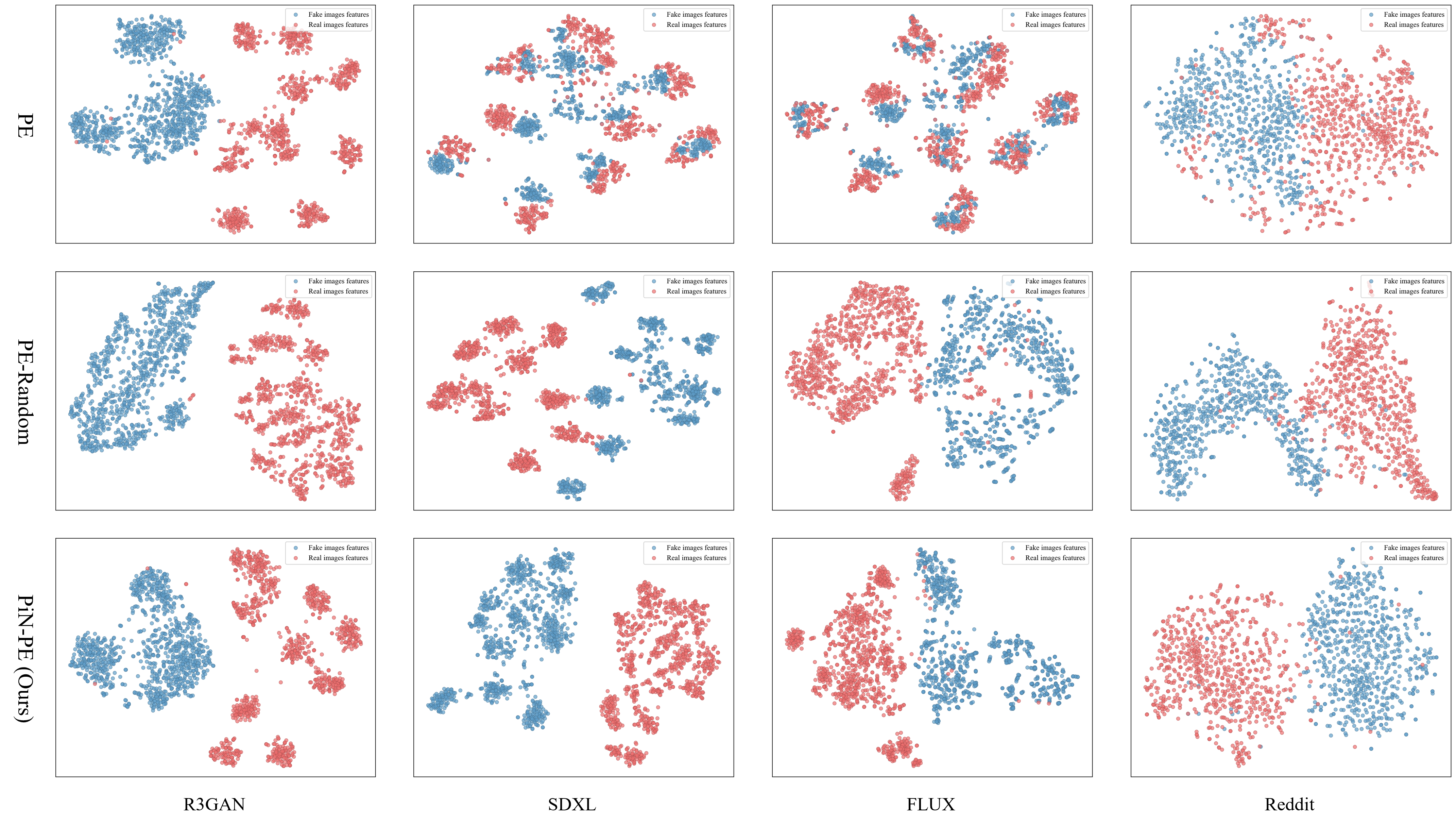}
    \caption{\textbf{T-SNE Visualization of Features Extracted Using PE, PE-Random and PiND (ours).} Our method achieves strong real/fake discrimination, while adding random noise to the visual features also improves discriminative ability. }
    \label{fig:Qualitative}
\end{figure*}

\noindent\textbf{Ablation Study of Different Backbones.} To assess the robustness of our method across backbone architectures, we evaluate it with multiple backbone networks, where the results are shown in Table \ref{tab:backbone}. Our method consistently improves the detection performance of all pre-trained models (the accuracy of PE-Core-B16-224 increases by 12\%), further validating that our Positive-incentive Noise effectively suppresses spurious shortcuts and enables the model to extract more generalized artifact representations. In addition, all selected backbones are ViT-based and are pre-trained on image-text pairs, providing a useful prior alignment between visual and textual representations that our method can leverage.

\noindent\textbf{Ablation Study of Hyperparameter $\lambda$.} In order to more comprehensively evaluate the effectiveness of our method, we also conduct ablation experiments on the hyperparameter $\lambda$, which can be shown in Table \ref{tab:lambda}. When our parameter $\lambda$ is set to 0.2 as set in the paper, the best results are achieved on all datasets. It is worth noting that overly large parameter values can degrade performance, which we attribute to excessive perturbation that overwhelms task-relevant cues and introduces harmful interference. 

\subsection{Robustness Evaluation}
In real-world scenarios, images are inevitably affected by unknown disturbances during transmission and interaction, which pose additional challenges for AI-generated image detection. To investigate robustness under such conditions, we further evaluate the performance of different detection methods against a variety of disturbances, such as JPEG Compression and Gaussian Blur (Quality Factor (QF) = 95, 75, 60) and Gaussian blur ($\sigma$ = 0.5, 1.5, 2.5). However, as noted in the AIGIBench\cite{li2025artificial}, when robustness evaluations are performed without robust data augmentation during training, most existing methods suffer a substantial performance drop to around 50\%, indicating limited robustness under such conditions. Accordingly, in our experiments, we retrained all methods and incorporated robust data augmentations during training, including random JPEG Compression ($\text{QF}\sim\text{Uniform} \; [30, 100)$) and random Gaussian Blur ($\sigma \sim \text{Uniform} \; [0.1, 3.0]$). Each augmentation is conducted with 10\% probability. Notably, our approach remains unchanged and does not incorporate any data augmentation operations. The results are shown in Table \ref{table:Robustness_GenImage} and Table \ref{table:Robustness_AIGIBench}. Despite not using any data augmentation, our method consistently outperforms others, maintaining a relatively higher accuracy in detecting AI-generated images. This indicates that our method has the potential to preserve strong detection performance under previously unseen perturbations.

\subsection{Qualitative Analysis}
To further assess the generalization ability of our method, we visualize the feature distributions before the binary classifier, as shown in Fig.~\ref{fig:Qualitative}. Owing to the strong representational capacity inherited from pre-training, the PE baseline already exhibits solid detection performance, particularly on the in-the-wild Reddit dataset, where real and fake samples are sometimes separable. Building on this, our method (PiND) further enhances generalization, maintaining a clear separation between real and fake categories even under previously unseen generators, which suggests that the proposed Positive-incentive Noise more effectively suppresses spurious shortcut cues and encourages the extraction of more stable, generator-agnostic artifact representations. Notably, compared with the original PE model, injecting random noise into the visual features also improves class separability, which further supports our insight that feature-level perturbations can strengthen the robustness and generalization of AI-generated image detectors.

\section{Conclusion}
In this paper, we systematically analyze the limitations of current AI-generated image detectors and identify spurious shortcuts as the root cause of overfitting. We are the first to explore how feature space perturbations affect the generalization of AI-generated image detectors and observe that introducing small random perturbations into the feature space can effectively mitigate the spurious shortcuts and significantly improve generalization. Building on this insight, we propose a novel framework, Positive-incentive Noise for AI-generated image detection(PiND), which transforms random noise perturbations into controllable, task-oriented signals. Specifically, we employ a cross-modal attention mechanism to integrate visual features and label embeddings, generating task-adaptive, forgery-conditional Gaussian perturbations. By jointly optimizing the noise generator and the visual encoder, PiND suppresses shortcut-sensitive directions under beneficial stochastic transformations while amplifying stable forensic cues, leading to more generalized artifact representations. Our method can be adapted to different ViT-based pre-trained models, opening a new direction for noise-driven learning in multimedia forensics.

\bibliographystyle{unsrt}
\bibliography{ieee}

\end{document}